\pdfoutput=1

\documentclass[11pt]{article}

\usepackage{authblk}
\usepackage[]{acl}

\usepackage{times}
\usepackage{latexsym}

\usepackage[T1]{fontenc}

\usepackage[utf8]{inputenc}

\usepackage{microtype}

\usepackage{graphicx}
\usepackage{amsmath}
\usepackage{cleveref}
\usepackage{xcolor}
\usepackage{algorithm2e}
\RestyleAlgo{ruled}
\usepackage[group-separator={,}]{siunitx}
\usepackage{booktabs}
\usepackage{tabularx}
\usepackage{hyperref}
\usepackage{mathtools}
\usepackage{enumitem}

\newcommand*\samethanks[1][\value{footnote}]{\footnotemark[#1]}

\title{BLIAM: Literature-based Data Synthesis for Synergistic Drug Combination Prediction}

\author[1,\Thanks{ Contributed equally}\space  ]{\bf Cai Yang}
\author[2,{\samethanks} \space]{\bf Addie Woicik}
\author[3]{\bf Hoifung Poon}
\author[2,\Thanks{ Corresponding author } ]{\bf Sheng Wang}
\affil[1]{School of Computing, Australian National University}
\affil[2]{Paul G. Allen School of Computer Science, University of Washington}
\affil[3]{Microsoft Research}
\affil[  ]{\texttt{swang@cs.washington.edu}}

\begin{document}
\maketitle
\begin{abstract}
Language models pre-trained on scientific literature corpora have substantially advanced scientific discovery by offering high-quality feature representations for downstream applications. However, these features are often not interpretable, and thus can reveal limited insights to domain experts. Instead of obtaining features from language models, we propose BLIAM, a literature-based data synthesis approach to directly generate training data points that are interpretable and model-agnostic to downstream applications. The key idea of BLIAM is to create prompts using existing training data and then use these prompts to synthesize new data points. BLIAM performs these two steps iteratively as new data points will define more informative prompts and new prompts will in turn synthesize more accurate data points. Notably, literature-based data augmentation might introduce data leakage since labels of test data points in downstream applications might have already been mentioned in the language model corpus. To prevent such leakage, we introduce GDSC-combo, a large-scale drug combination discovery dataset that was published after the biomedical language model was trained. We found that BLIAM substantially outperforms a non-augmented approach and manual prompting in this rigorous data split setting. BLIAM can be further used to synthesize data points for novel drugs and cell lines that were not even measured in biomedical experiments. In addition to the promising prediction performance, the data points synthesized by BLIAM are interpretable and model-agnostic, enabling \textit{in silico} augmentation for \textit{in vitro} experiments.

\end{abstract}

\section{Introduction}
\label{sec:intro}
Scientific literature is an important resource to provide prior knowledge for scientific discovery \cite{Lu_Wang2020-lp, Wei2013PubTatorAW}. Natural language processing techniques have been extensively developed to mine scientific literature for scientific problems, including material property prediction \cite{Tshitoyan2019UnsupervisedWE}, COVID therapeutics research \cite{Lever2021AnalyzingTV}, cancer research \cite{Lever2019-ur}, and drug repurposing \cite{Hsiao2019-bc, Detroja2022-ga}. A promising line of research is to train domain-specific language models on biomedical corpora, which have been used to produce dense representations for downstream supervised learning tasks \cite{Gu2021-qf, Luo2022-hx, Lee2019-gm}.

However, these dense representations are not interpretable to domain experts, which is critical for biomedical applications \cite{Stiglic2020-qj}. Although existing interpretable methods can highlight important features within the embeddings \cite{Lundberg2017-wg, Shrikumar2017-eb}, these features are still latent dimensions. Here, we propose to directly synthesize new labeled biological datapoints using pretrained LMs (PLMs). The synthesized data points will be used as augmentations to help train the downstream classifier. A critical issue of synthesizing new data points is the potential data leakage, where test data points might be explicitly mentioned in the scientific literature and thus be seen by PLMs. To address this issue, we introduce a novel large-scale drug combination discovery dataset GDSC-combo \cite{Jaaks2022-xs} that was published in April 2022 after the corpus that the biomedical PLM used for training \cite{Gu2021-qf}. By examining this dataset in 18 million PubMed abstracts that are seen by the PLM, we found that 99.99\% of data points in GDSC-combo never appeared in any abstract. Here, the task is to predict whether a triplet of two drugs and a cell line will be synergistic or not.


\begin{figure*}[!ht]
  \centering
    \includegraphics[width=0.99\textwidth]{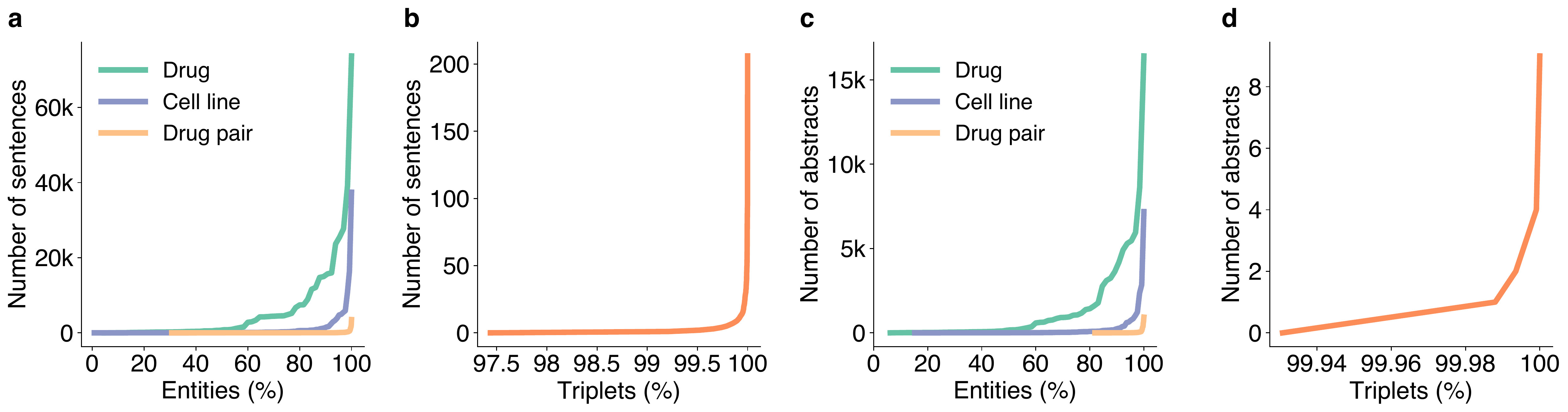}
  \caption{\textbf{Examining data leakage by counting the number of sentences and abstracts that mention drugs and cell lines.} Sentences and abstract are obtained from \num{34592344} PubMed papers. x-axis shows the percentage of entities (\textbf{a, c}) or triplets (\textbf{b, d}) appear in fewer than $k$ sentences (\textbf{a, b}) or abstracts (\textbf{c, d}), where $k$ is shown in y-axis. Drug pair stands for two drugs. Triplet stands for two drugs and a cell line.}
  \label{fig:leak}
\end{figure*}

We propose the Biomedical Literature Iterative Augmenting Module (BLIAM) to synthesize new triplets from the PLM, which we in turn use to augment the GDSC-combo dataset. BLIAM first mentions sentences that mention two drugs and a cell line in our vocabulary list based on the original dataset \cite{Jaaks2022-xs}. It then clusters these sentences and derives representative prompts according to cluster centers. These prompts are then used to synthesize new triplets based on the PLM. BLIAM iteratively performs these three steps to create a large and high-quality synthesized synergistic dataset, which is later combined with the original dataset \cite{Jaaks2022-xs} to train a downstream classifier. 

We found that augmenting the GDSC-combo dataset substantially improved downstream synergy prediction from 0.30 AUPRC to 0.38 AUPRC compared to a non-augmentation approach, and that the amount of improvement grows with repeated iterations of BLIAM. BLIAM was particularly beneficial when applied to never-before-seen drugs and cell lines at test time, representing a more realistic setting for real-world clinical applications. In summary, we have introduced a new dataset and task for literature-based data augmentation that does not have the data leakage issue. We have also developed a literature-based data synthesis approach that successfully generates high-quality training data points.


\begin{figure*}[!ht]
  \centering
    \includegraphics[width=0.99\textwidth]{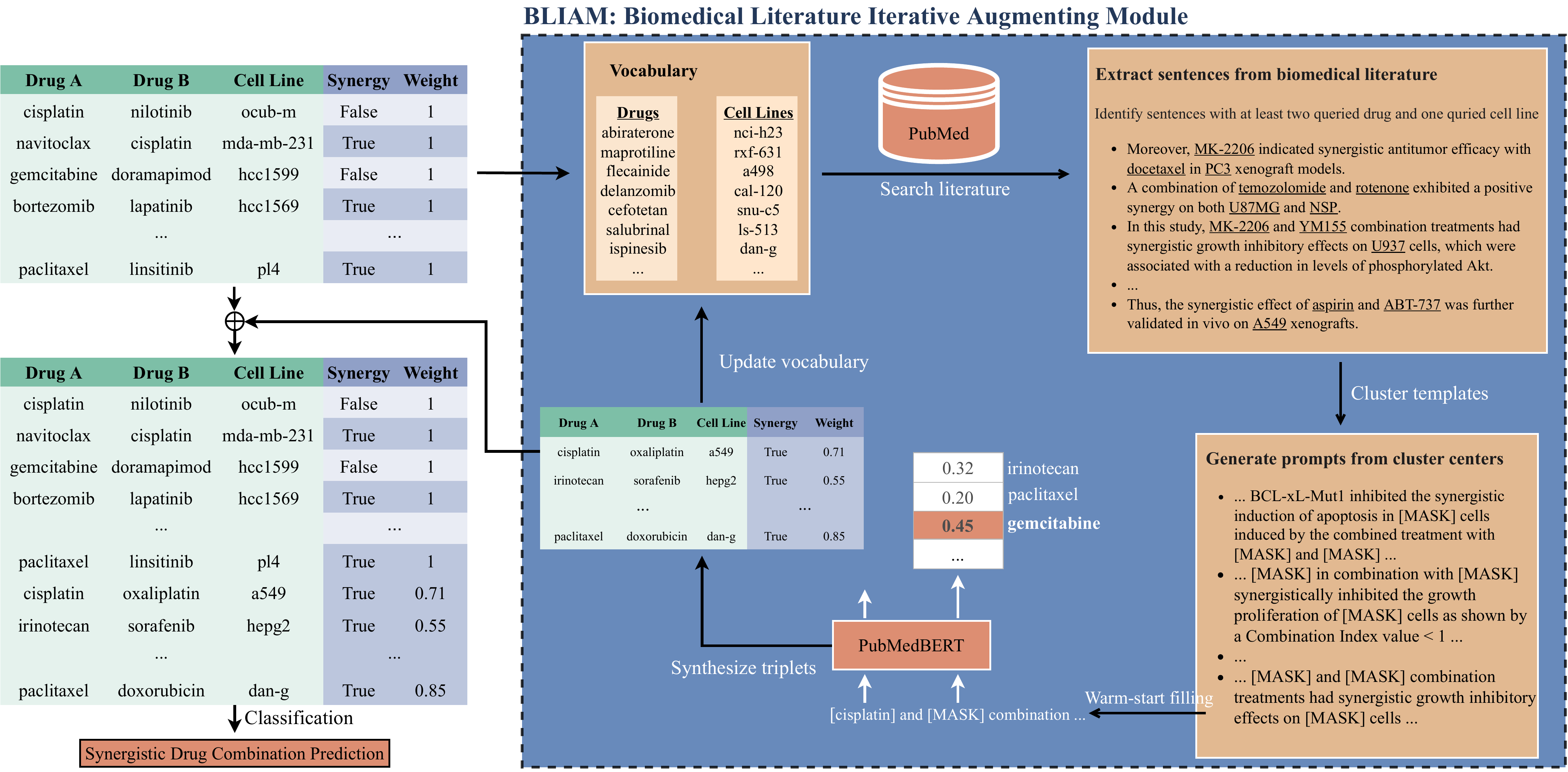}
  \caption{\textbf{Overview of BLIAM.} BLIAM begins by searching PubMed literature using an initial vocabulary of drugs and cell lines. The sentences obtained are then clustered and medoids of each cluster are used to create prompts for PubMedBERT for cloze prompt filling. Tokens filled in the drug and cell line masks are used to expand the initial vocabulary, even if they are not valid drug and cell line names. These steps (blue box) are performed iteratively. After 3 iterations, generated triplets are used as an augmentation dataset for synergy prediction task.}
  \label{fig:process}
\end{figure*}

\section{Task and Dataset}
\label{sec:data}

We focus on drug synergy classification, an important biomedical task to predict whether the combination of two drugs will have a greater-than-additive drug response effect on a given cell line \cite{Jaaks2022-xs}.  Predictive models can help fill a large void for synergy screens, as it is too expensive and time-consuming to exhaustively search the combinations of drugs and cell lines \textit{in vitro}. However, the sparsity of available labeled drug synergy datasets also makes this a challenging problem, and effectively leveraging prior knowledge through existing literature represents one possible route toward improved synergy prediction.

We formulate the problem of drug synergy prediction as a binary classification task. We obtained an initial set of input triplets and corresponding synergy labels $\mathcal{D}_{full} = \{ ((d^A_i, d^B_i, c_i), y_i) \}_{i=1}^{N'}$ from a recent dataset \cite{Jaaks2022-xs}, where $y_i \in \{0, 1\}$ indicates whether drugs $d^A_i$ and $d^B_i$ are synergistic in cell line $c_i$. Based on this initial set, we could then learn the classifier $f_\theta(d^A_i, d^B_i, c_i) \in [0, 1]$, which reflects the probability that a triplet is synergistic.

We further filtered the dataset $\mathcal{D}_{full}$ using the PubMedBERT \cite{Gu2021-qf} vocabulary $\mathcal{V}$, where we only retained triplets with $d^A_i \in \mathcal{V} \lor d^B_i \in \mathcal{V} \lor c_i \in \mathcal{V}$, producing a reduced dataset $\mathcal{D}$ of size $N = \num{24890}$, where only $191$ triplets have synergistic effect $y_i = 1$. In total, $\mathcal{D}$ contains $125$ unique cell lines and $65$ unique drugs.  In this work, we do not consider additional classification features for drugs or cell lines, such as gene expression or drug SMILES representation. However, our framework can also incorporate such features as long as we can obtain these features from other databases after we have synthesized the triplets.

Our goal is to use the pretrained PubMedBERT \cite{Gu2021-qf} to generate a synthetic dataset $\Tilde{\mathcal{D}}$ and learn the classifier $f_\theta(d^A_i, d^B_i, c_i) \in [0, 1]$ using $\Tilde{\mathcal{D}} \cup \mathcal{D}$. Importantly, the drug synergy dataset from \citeauthor{Jaaks2022-xs} was published after the PLM from \citeauthor{Gu2021-qf}, enabling us to avoid data leakage between the PLM and the synergy classification data splits. \textbf{Fig. \ref{fig:leak}} shows the proportion of entities and their appearance in sentences and abstracts. Despite single drugs, single cell lines, and drug pairs frequently appearing in sentences and abstracts, triplets exhibit low occurrence in PubMed literature, further confirming that using PubMedBERT to augment GDSC-combo does not bring the risk of data leakage.




\section{Methods}
\label{sec:method}

\subsection{Base classifier}
For a given triplet $(d^A_i, d^B_i, c_i) \in \mathcal{D}$, we aim to learn a classifier $f_\theta(d^A_i, d^B_i, c_i) \rightarrow [0, 1]$. In particular, we design a classifier consisting of two embedding matrices for drugs and cell lines. We didn't use additional features for drugs or cell lines so the embedding matrices are randomly initialized. The embedding vectors for three entities from the triplet are concatenated and passed into a multi-layer neural network with Leaky ReLU activation functions for classification. We followed previous work on drug synergy prediction for the neural network architecture \cite{Preuer2018-vm}. 

When augmented data is available, we combine the drug synergy dataset $\mathcal{D} = \{((d^A_i, d^B_i, c_i), y_i) \}_{i=1}^N$ with the augmented synergistic dataset $\Tilde{\mathcal{D}} = \{((\tilde{d^A_j}, \tilde{d^B_j}, \tilde{c_j}), 1)\}_{j=1}^M$ and learn a new classifier $f^{'}_\theta(d^A, d^B, c) \in [0, 1]$. We train the model with binary cross entropy as
$\mathcal{L}(\theta | \mathcal{D}, \Tilde{\mathcal{D}}) = \sum_{i=1}^N \texttt{BCE}(y_i, f^{'}_\theta(d^A_i, d^B_i, c_i)) +   \sum_{j=1}^M w_j \log(f^{'}_\theta(\Tilde{d^A_j}, \Tilde{d^B_j}, \Tilde{c_j}))$,
where \texttt{BCE} indicates binary cross entropy loss and $w_j$ is a weight factor according to the likelihood of each synthetic triplet generated by PLM. At inference time, we predict a triplet to be synergistic if $f^{'}_\theta(d^A, d^B, c) > 0.5$.

\subsection{Manual Prompts}
We first develop a few manual prompts to augment the data. We manually construct a list of cloze prompts that can be filled by PubMedBERT (see \textbf{Supplementary \ref{app:man}}), such as {\fontfamily{cmvtt}\selectfont On cell line [MASK], [MASK] has synergy with [MASK]} and {\fontfamily{cmvtt}\selectfont [MASK] and [MASK] are effective to treat cell line [MASK]}. We then take the most likely token output by the PLM for each masked token to construct new drug combinations and cell line triplets, to which we assign positive $y=1$ synergy labels. We denote the union of the original dataset and this simulated dataset as $\Tilde{\mathcal{D}}_{man}$. 

\subsection{Literature-mined prompts}

Although manual prompts are more intuitive, designing good templates requires domain knowledge and may not cover all aspects of a specific task. To address this, we extract sentences mentioning drugs and cell lines from scientific literature, which serve as candidates for automated prompt generation (\textbf{Fig. \ref{fig:process}}).

To search for candidate templates, we first build a larger vocabulary of single drugs and single cell lines by combining the set of all drugs and cell lines from our dataset $\mathcal{D}$ with the drugs and cell lines from some existing databases, including LINCS \cite{Subramanian2017-an}, Genomics of Drug Sensitivity in Cancer (GDSC) \cite{Iorio2016-ny}, Cancer Cell Line Encyclopedia (CCLE) \cite{Meyers2017-gy, Tsherniak2017-jz}, and National Cancer Institute 60 human cell line (NCI-60) \cite{Shoemaker2006-nx}. This increases the number of drugs from \num{125} to \num{1486} and the number of cell lines from \num{65} to \num{2037}, ensuring that our collection of drugs and cell lines is large enough to identify diverse templates. This step does not introduce data leakage as these external databases are not drug combination databases and we only collect drug and cell line names from them.

We then search within the PubMed abstract collection \cite{Wei2019-gj} for sentences containing at least two drugs and one cell line, which can therefore be interpreted as a triplet. After obtaining this set of sentences, we then filter them by keywords such as \textit{synergy} and \textit{synergistic} to ensure that the sentence relates to synergy.

We can then mask the triplets within each sentence to obtain a prompt. However, as treating every resulting sentence as a prompt would be time-consuming and redundant, we only retain representative prompts. Specifically, we masked all drug and cell line mentioned within a sentence and computed its BERT embedding. Then, we compute the k-medoid clustering of the masked sentence embeddings. Finally, we selected the medoid of each cluster to use as the cluster's representative prompt.

\begin{figure*}[!ht]
  \centering
    \includegraphics[width=0.99\textwidth]{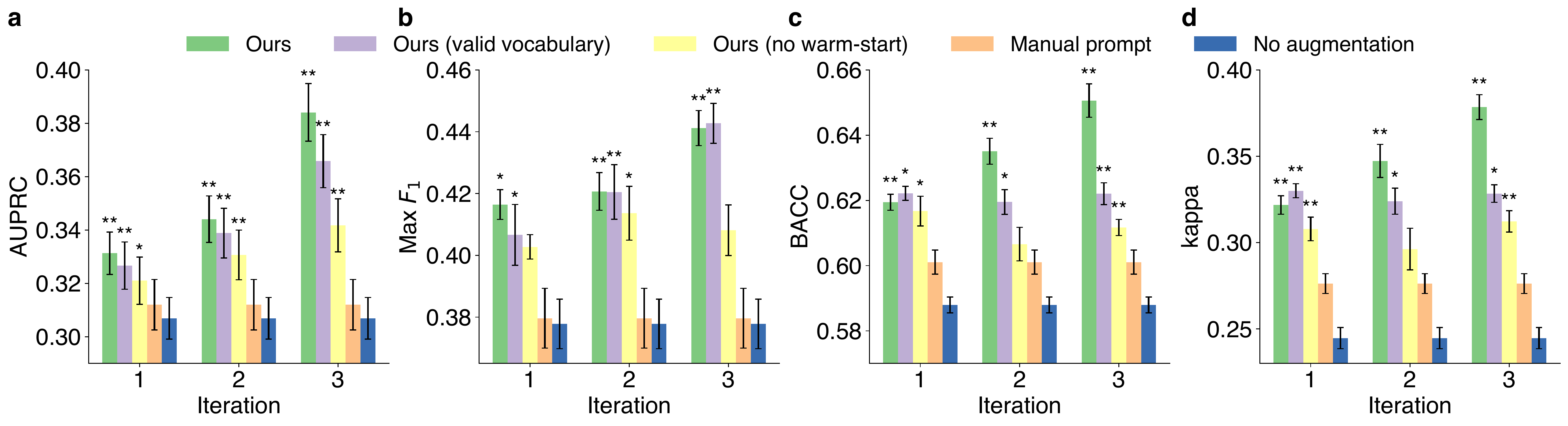}
  \caption{\textbf{Performance on downstream drug synergistic prediction evaluated using four metrics.} Vertical bars represent standard error bars. Stars above bars indicate statistical significance when comparing with \textit{No augmentation} setting using one-sided paired t-tests ($*$: $p < 0.05$, $**$: $p < 0.01$). y-axis shows the iteration. Ours, Ours (valid vocabulary), Manual prompt, No augmentation stand for training on $\mathcal{\Tilde{D}}_{it},\mathcal{\Tilde{D}}_{valid},\mathcal{\Tilde{D}}_{manul}$, and $\mathcal{D}$, respectively. Ours (no warm-start) stands for our methods without warm-start filling. }
  \label{fig:overall_performance}
\end{figure*}
\subsection{Prompt warm-start filling}
We found that directly filling in three masked tokens in a prompt could generate many redundant triplets, even using different prompts, thus leading to less efficient generation. To address this issue, we have developed a warm-start filling strategy. Specifically, we uniformly sample a triplet from the current dataset and uses one or two elements in this triplet to pre-fill the prompt. This strategy will avoid frequently filling the prompt with popular drugs or cell lines, thus enabling us to synthesize triplets containing rare cell line or drugs and improving downstream classification results.

Since some triplets might contain drug or cell line name that is not in the PubMedBERT vocabulary, we further restrict to only warm-start fill in-vocabulary entity. For instance, given the triplet {\fontfamily{cmvtt}\selectfont (Cisplatin, Camptothecin, BT-483)} and prompt template {\fontfamily{cmvtt}\selectfont [MASK] and [MASK] combination treatments had synergistic growth inhibitory effects on [MASK] cells}, Cisplatin is the only triplet element that appears in the PubMedBERT vocabulary. Therefore, we could pre-fill the prompt template as {\fontfamily{cmvtt}\selectfont  Cisplatin and [MASK] combination treatments had synergistic growth inhibitory effects on [MASK] cells}, or {\fontfamily{cmvtt}\selectfont  [MASK] and cisplatin combination treatments had synergistic growth inhibitory effects on [MASK] cells}. All token filling are type-aware: a drug (cell line) mask  will only be replaced with a drug (cell line) name. In the case where the prompt template contains additional {\fontfamily{cmvtt}\selectfont [MASK]} tokens (i.e., more than two drug mask positions or more than one cell line mask position), all masked locations are considered possible options for warm-start filling.

We then use these warm-start filled prompts, all containing at least one remaining masked token, to further prompt the PLM. In the case of more than three {\fontfamily{cmvtt}\selectfont [MASK]} tokens, we enumerate all combinations of two drugs and one cell line to form new triplets. This warm-start filling strategy helps us achieve better diversity while still consider contextual information.

\subsection{Iterative augmentation} \label{sect:iterative_aug}

Furthermore, we repeat the literature-mined prompting process multiple times to iteratively construct the augmented dataset (\textbf{Fig. \ref{fig:process}}). Iteratively adding new triplets to the dataset enables BLIAM to expand its set of possible prompts based on previous literature-based findings. This can both increase the potential diversity of prompts and improve prompting quality, with subsequent iterations able to learn from the synthesized data triplets from previous iterations. Given the initial dataset $\mathcal{D}$, we fill literature-mined prompts to produce a new triplet dataset $\Tilde{\mathcal{D}}_{liter}$. We then mine literature again, including triplets from $\mathcal{D} \cup \Tilde{\mathcal{D}}_{liter}$, fill the resulting prompts, and query PubMedBERT to produce the newly augmented triplet dataset $\Tilde{\mathcal{D}}_{liter}^{(2)}$. We repeat this process for a total of $\gamma$ iterations, using the simulated triplets from the previous iteration to help inform the prompts for the next, and denote the resulting iterated dataset as $\Tilde{\mathcal{D}}_{it} = \mathcal{D} \cup \mathcal{\Tilde{D}}_{liter}^{(3)}$.

\subsection{Vocabulary-restricted augmentation}
Although we provide context to PubMedBERT in the form of the filled-in prompts, it is possible that the most likely generated words for the {\fontfamily{cmvtt}\selectfont [MASK]} tokens do not correspond to actual drugs or cell lines. We address this in one variant of our method, where we restrict the set of generated tokens we consider to be within the set of drugs or cell lines from $\mathcal{D}$, LINCS, GDSC, CCLE, and NCI-60 for drug and cell line mask masks, respectively. This ensures that the generated triplets do in fact correspond to two drugs and a cell line. This additional filtering step makes the synthesized triplets and consequent classifier performance more interpretable and also enables biologists to inspect and understand BLIAM's synthesized dataset. We then repeat the iterative process from Section \ref{sect:iterative_aug} and refer to the resulting iterated dataset, now restricted to valid triplets, as $\Tilde{\mathcal{D}}_{valid}$.



\begin{figure*}[!ht]
  \centering
    \includegraphics[width=0.99\textwidth]{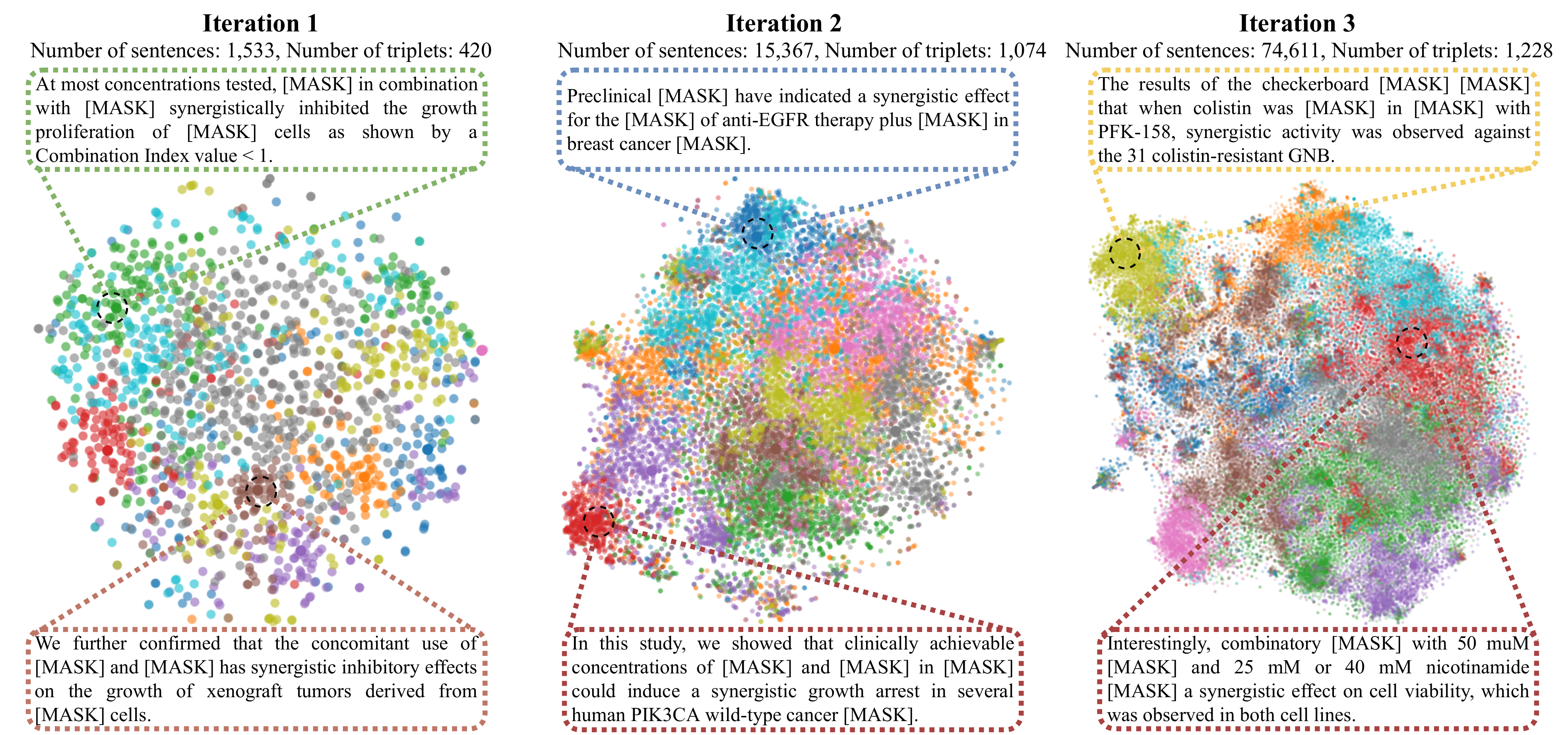}
  \caption{\textbf{t-SNE plots on literature-mined sentences in three iterations.} Sentences are clustered into ten clusters in each iteration. Each sentence is colored according to its belonging cluster. One prompt template is derived according to the center of each cluster. }
  \label{fig:tsne}
\end{figure*}

\section{Experimental setting}
\label{sec:exp}

We chose PubMedBERT-abstract\footnote{\url{https://huggingface.co/microsoft/BiomedNLP-PubMedBERT-base-uncased-abstract}, Last accessed: December 11, 2022.} as our PLM. After intersecting the dataset with PubMedBERT's vocabulary, there are \num{24890} instances, 191 of which have a positive label. We repeated the recursive search in literature-mined prompts for $\gamma = 3$ iterations. 

We conducted the triplet synergy classification task through 5-fold stratified cross-validation. We trained the classifier using the Adam optimizer with $\epsilon=10^{-8}, \beta_1 = 0.9, \beta_2 = 0.999$ for $50$ epochs. We set the batch size to $64$. We performed grid search on learning rates over $[0.01, 0.005, 0.001]$, hidden dimensions over $[128, 256, 512]$, and warm-up epochs over $[0, 5, 10, 20, 30, 40]$. The maximum number of parameters of the classifier is \num{343874}. For training with the augmentation dataset, we considered whether using instance weights on loss from synthetic instances as an additional hyperparameter. All experiments were conducted on Nvidia RTX 3090 GPU with 24GB Memory. Training and inference can be finished within one GPU hour. We followed previous works to adopt four metrics that can evaluate the imbalanced classification task \cite{Lin2022-ri}: AUPRC, max F1 score, balanced accuracy (BACC), and Cohen's kappa (kappa). 

\section{Experimental results}
\label{sec:result}

\subsection{BLIAM-derived augmentations improve downstream classification}

We first investigated whether the BLIAM-derived augmentations can improve downstream drug synergy classification. We compared the performance of training using the unaugmented dataset $\mathcal{D}$ with the three variants of PLM-augmented datasets: $\Tilde{\mathcal{D}}_{it}$ and $\Tilde{\mathcal{D}}_{valid}$ produced by BLIAM, and $\Tilde{\mathcal{D}}_{man}$ from the manual prompting. We found that all three augmented datasets outperformed the unaugmented dataset $\mathcal{D}$ (\textbf{Fig. \ref{fig:overall_performance}}), indicating the effectiveness of using literature to generate high-quality biomedical training data points. 

Among these three augmented datasets, the $\Tilde{\mathcal{D}}_{it}$ augmented dataset led to the best overall performance, with an AUPRC of 0.38, max $\text{F}_1$ score of 0.44, BACC of 0.65, and kappa of 0.38, relative to the 0.31, 0.38, 0.59, and 0.25 respectively for the unaugmented $\mathcal{D}$. $\Tilde{\mathcal{D}}_{valid}$ restricts the prompts to only synthesize valid triplets of drugs and cell lines, thus providing interpretable triplets for domain experts. We found that the performance of training on $\Tilde{\mathcal{D}}_{valid}$ is slightly worse than $\Tilde{\mathcal{D}}_{it}$, suggesting a trade-off between restricted interpretable results and downstream prediction performance. Nevertheless, the performance of $\Tilde{\mathcal{D}}_{valid}$ is still better than $\Tilde{\mathcal{D}}_{man}$ and  $\mathcal{D}$, again confirming the benefit of our iterative augmentation approach. We also observed that our method outperformed the variant with warm-start, demonstrating the effectiveness of warm-start in prompt filling.

Moreover, we noticed that the performance of $\Tilde{\mathcal{D}}_{it}$ improves over iterations. For example, the AUPRC of  $\Tilde{\mathcal{D}}_{it}$ improves from $0.33$ to $0.34$ after the second iteration and further boosts to $0.38$ after the third iteration. Since our approach generates new prompts according to the data points synthesized at each step, these new prompts are learned using a larger data set and thus might be more robust. We did not observe further improvement after more than three iterations. Collectively, the improvement of our approach against the unaugmented approach and manual prompting demonstrates the preeminence of iteratively using a PLM to synthesize new data points.

\begin{figure*}[!ht]
  \centering
    \includegraphics[width=0.99\textwidth]{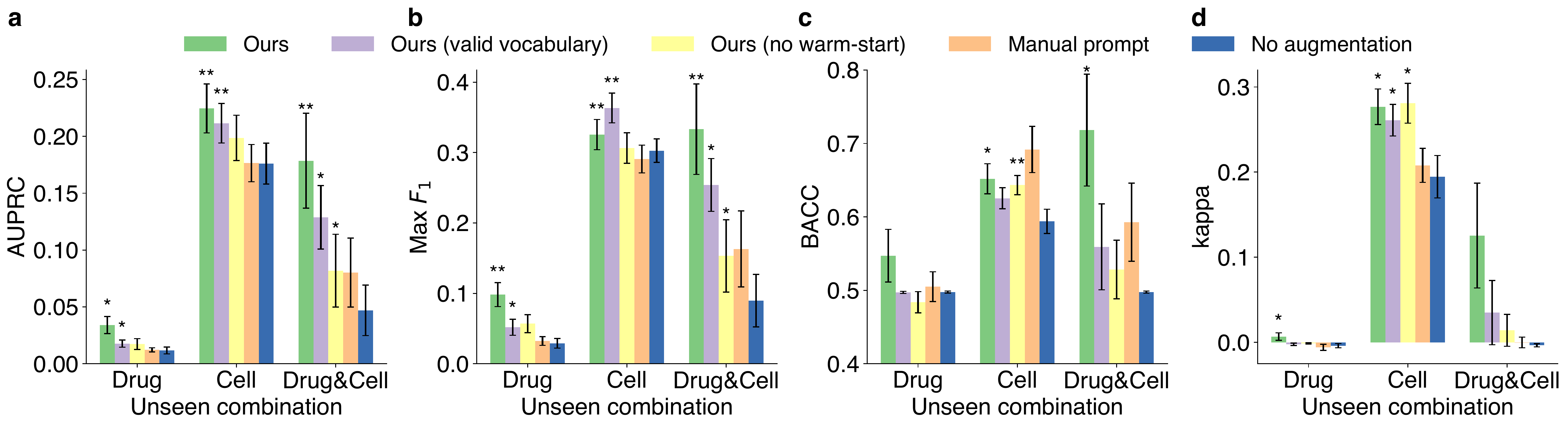}
  \caption{\textbf{Performance on triplets that contain never-before-seen drugs or cell lines evaluated using four metrics.} Vertical bars represent standard error bars. Stars above bars indicate statistical significance when comparing with \textit{No augmentation} setting using one-sided paired t-tests ($*$: $p < 0.05$, $**$: $p < 0.01$). Ours, Ours (valid vocabulary), Manual prompt, No augmentation stand for training on, $\mathcal{\Tilde{D}}_{it},\mathcal{\Tilde{D}}_{valid},\mathcal{\Tilde{D}}_{manul}$, and $\mathcal{D}$, respectively. Ours (no warm-start) stands for our methods without warm-start filling. Performance of iteration 3 is shown here.}
  \label{fig:unseen_performance}
\end{figure*}

\subsection{Iterative augmentation expands and improves the augmented triplets}

To further investigate the benefit of gradually expanding the augmented dataset with repeated iterations of the augmenting module, we inspected the prompts identified at each of our three iterations. Manual prompts identified a total of \num{636} simulated triplets. By contrast, the first iteration of literature-based prompting led to \num{420} simulated triplets, which increased to \num{1074} and \num{1228} simulated triplets in the second and third iterations, respectively, suggesting larger and more robust augmentations. These triplets were also generated from an increasing number of sentences (\textbf{Fig. \ref{fig:tsne}}). We further examined the t-Stochastic Neighbor Embedding (t-SNE) \cite{Maaten2008-ed} representations of sentences extracted at each iteration (\textbf{Fig. \ref{fig:tsne}}). Here, the prompt template is derived from the center of each cluster. As the set of prompt templates expands over iterations, we found that the clustering pattern of the sentences becomes more visible. This in turn suggests that the clustering for prompt template selection becomes more robust at later iterations of the augmenter. Interestingly, we found that the prompt templates at later iterations are less compatible with human intuition. For example, despite the existence of drug-related context, templates in the third iteration do not seem to mask a drug or cell line token. This observation is consistent with previous works \cite{Shin2020-ou} where automatically derived prompts often contain non-informative tokens. Nevertheless, although the prompt templates might not be intuitive to human experts, the generated triplets are interpretable and can help biologists identify new synergistic triplets.

\subsection{BLIAM synthesizes triplets that contain never-before-seen drugs and cell lines}

A critical limitation of experimentally-derived drug synergistic datasets is the existence of new drugs and cell lines at inference time, which hinders the progress of  predicting patient outcomes in real-world clinics. Incorporating prior knowledge from biomedical literature might help to address this problem. To investigate our ability to leverage the PLM to classify triplets that contain never-before-seen drugs and cell lines, we designed three new test settings: \textit{Drug}, \textit{Cell}, and \textit{Drug\&Cell}. Given a set of test triplets $\{(d^A_i, d^B_i, c_i)\}_{i=1}^T$, we define each setting as follows: in \textit{Drug}, we exclude any triplets containing $d^A_i$ or $d^B_i$ from the input training dataset $\mathcal{D}$ for all $i$; in \textit{Cell}, we exclude any triplets containing $c_i$ from $\mathcal{D}$ for all $i$; in \textit{Drug\&Cell}, we exclude any triplets containing $d^A_i$, $d^B_i$, or $c_i$ from $\mathcal{D}$ for all $i$.

In these challenging but more realistic settings, we found that BLIAM was again beneficial for synergistic classification, relative to no augmentation or manual prompting (\textbf{Fig. \ref{fig:unseen_performance}}). For example, in \textit{Cell}, the AUPRC was 0.18 without augmentation or with manual prompting, and increased to 0.22 with BLIAM's augmented dataset. The AUPRC was also better if we restricted the vocabulary (0.21) or even skipped the warm-start filling (0.20). The improvement was even larger on the most challenging setting of \textit{Drug\&Cell}, where the unaugmented $\mathcal{D}$ had an AUPRC of 0.05, $\Tilde{\mathcal{D}}_{man}$ had an AUPRC of 0.08, and our $\Tilde{\mathcal{D}}_{it}$ had an AUPRC of 0.18. While \textit{in vitro} experimental approaches might be too expensive or otherwise infeasible to measure certain drugs and cell lines, BLIAM complements them by synthesizing high-quality training data points, paving the way for applying these experimental measurements to real-world clinics.

\section{Related Work}
\label{sec:related}

\subsection{PLMs for data augmentation}
Recent works have shown that large PLMs can function as good few-shot and zero-shot learners \cite{Brown2020-mo, Radford2019-ud}, and thoughtful probing of PLMs has been used to improve downstream model performance in low-data settings \cite{Zhong2021-hf}. PLMs have further been used for data augmentation, including word-substitution-based approaches \cite{Hu2019-ge}, sentence-substitution \cite{Kumar2019-ai}, backtranslation \cite{Xie2020-ij}, prompting \cite{Wang2022-qh, Shin2020-ou, Schick2021-hy}, and textual generation for synthetic dataset construction \cite{Yang2020-jh, Mekala2022-mq, Zhou2022-ws, Puri2020-pz, Vu2021-rk, meng2022generating, Ye2022-ca, Yoo2021-em, Lewis2019-qa, Wei2021-dd}. In particular, PLM-based augmentation techniques have improved downstream model performance in imbalanced classification settings \cite{Liu2020-lq}, where generating synthetic samples for the minority class improves performance \cite{Liu2020-lq, Chawla2002-qo}.

Most similar to this work, several data augmentation techniques use cloze test formulations \cite{Chapelle1990-sm}, where the PLM is used in a fill-in-the-blank task \cite{Ng2020-rj, Cai2020-yw, Schick2021-hy} to augment a textual corpus. In this work, however, we use a cloze test formulation to synthesize novel, interpretable data points for downstream biomedical applications. The generated tokens are therefore taken out of their surrounding textual context after the iterative dataset synthesis is complete. Moreover, we prevent data leakage by carefully designing our task.

\subsection{PLMs as relational extractors}
PLMs have also been shown to function as good relational extractors, effectively distilling information contained in the pretraining textual corpus even in the few-shot setting \cite{Brown2020-mo}. Several works have found that using cloze sentence formulations with PLMs can behave as relation extractors \cite{Petroni2019-ad, Jiang2020-fz}, presenting an alternative to large knowledge bases \cite{Baldini_Soares2019-eg, Shwartz2020-tw}. Most similar to our work, \citeauthor{Bouraoui2020-to} use a known relational pair to identify relevant templates and use a PLM to predict other tokens that may belong to the same template relation. In this work, however, we study how literature and PLM can improve downstream application that does not involve text data. 

\subsection{Literature mining for scientific discovery}
NLP and literature-mining techniques have proven to be a useful technique to digest vast amounts of scientific and medical text and extract useful knowledge. Downstream applications include material science property prediction \cite{Tshitoyan2019-kx}, clinical outcome forecasting \cite{Naik2022-rl, Van_Aken2021-im, Jin2021-wf}, yeast cell-signaling modeling \cite{Coutant2019-go}, cancer gene identification \cite{Hsiao2019-bc, Lever2019-ur}, and drug repurposing \cite{Nye2021-uc, Jin2021-wf, Tworowski2021-bg}. One work uses PLMs for a biomedical knowledge graph (KG) completion task \cite{Nadkarni2021-yz}, though does not focus on downstream applications of such KGs as what we did. Recently, \citeauthor{Shim2022-pb} employed document-based feature extraction to improve anti-cancer drug synergy prediction performance \cite{Shim2022-pb}. In this work, we build upon these ideas and use a PLM to synthesize interpretable triplets for dataset augmentation.

\section{Conclusion and Future work}
\label{sec:discuss}
In this work, we have demonstrated that iteratively prompting PLMs with literature-mined templates produces interpretable data points that improve performance on a downstream drug synergy prediction task. We found that BLIAM's augmented dataset outperforms no augmentation and manual prompting, even though the prompts themselves are not always readily interpretable. We further found that BLIAM-synthesized triplets were particularly useful for drugs and cell lines not present in the original biomedical dataset, but which might have some literature evidence distilled by the PLM.

Although literature-mined prompts for PLMs improve our drug synergy classification performance, there are some direct extensions to the methodology that could further improve and refine the augmented dataset. First, when searching for sentences that contain two drugs and a cell line, we do not consider the overall intent of the sentence. For instance, we do not investigate negation before assigning a positive synergy label (e.g., {\fontfamily{cmvtt}\selectfont [MASK] and [MASK] are \emph{not} synergistic in cell line [MASK]}), or general research statements that are not intended to convey a result (e.g., {\fontfamily{cmvtt}\selectfont In this work, we investigated the possible synergy of [MASK] and [MASK] in [MASK] models}). Future work could therefore combine a sentiment analysis approach with our prompt construction to identify templates that truly imply a discovered synergistic triplet. Second, we limit the diversity of our generated triplets by always selecting the most likely generated word at each of our masked tokens. Instead, future work could expand the number of generated triplets by sampling from the generated distribution and continuing to weight the augmented dataset according to triplet likelihood in the classification loss to minimize the impact of false positives.

Furthermore, LMs pretrained on biomedical corpora could be applied more directly to biomedical tasks. Rather than constructing an augmented triplet dataset for drug synergy prediction, a PLM could be directly queried with {\fontfamily{cmvtt}\selectfont Cisplatin and [MASK] are synergistic in cell line dan-g} to produce a distribution over possible secondary drugs and directly answer the drug synergy question. Moreover, such approaches can be combined with recent prompting techniques such as chain-of-thought prompting \cite{Wei2022-zd}. Building upon recent work \cite{Liang2022-nw}, future studies could also combine other modalities to convey additional relevant biomedical information to the PLM, such as the molecular structure of a drug or marker genes for the cell line.

\section{Limitations}
Although we found that BLIAM improved downstream drug synergy prediction, there are several important limitations. First, the prompt template quality is dependent on the literature sentences containing the biomedical entities of interest. Second, the iterative prompting procedure could amplify inaccurate triplets. Third, the conditions measured in biomedical datasets and literature may contain their own biases, which are in turn internalized by the PLM and amplified by the iterative prompter. Finally, the diversity of generated triplets is restricted to PLM used. Although we can restrict the output to valid drugs and cell lines, it is unknown whether the drug pairs from the synthetic triplet are indeed synergistic on the cell line.

\bibliography{bib}
\bibliographystyle{acl_natbib}

\appendix

\section{Appendix}
\label{sec:appendix}

\subsection{Manual Prompts}
\label{app:man}
\begin{itemize}[leftmargin=0.25cm]
    \setlength\itemsep{0cm}
    \item {\fontfamily{cmvtt}\selectfont On cell line [MASK], [MASK] has synergy with [MASK].}
    \item {\fontfamily{cmvtt}\selectfont On cell line [MASK], [MASK] are synergistic with [MASK].}
    \item {\fontfamily{cmvtt}\selectfont [MASK] has synergy with [MASK] on cell line [MASK].}
    \item {\fontfamily{cmvtt}\selectfont [MASK] and [MASK] are synergistic on cell line [MASK].}
    \item {\fontfamily{cmvtt}\selectfont On cell line [MASK], there is a synergy between [MASK] and [MASK].}
    \item {\fontfamily{cmvtt}\selectfont There is a synergy between [MASK] and [MASK] on cell line [MASK].}
    \item {\fontfamily{cmvtt}\selectfont [MASK] and [MASK] are effective to treat to cell line [MASK].}
    \item {\fontfamily{cmvtt}\selectfont [MASK] and [MASK] are effective on cell line [MASK].}
    \item {\fontfamily{cmvtt}\selectfont On cell line [MASK], [MASK] and [MASK] are effective.}
    \item {\fontfamily{cmvtt}\selectfont On cell line [MASK], [MASK] and [MASK] are synergistic.}
    \item {\fontfamily{cmvtt}\selectfont On cell line [MASK], [MASK] and [MASK] have an synergy.}
\end{itemize}

\subsection{Literature-mined prompts}
\label{app:lit}

\subsubsection{Iteration 1}
\begin{itemize}[leftmargin=0.25cm]
    \setlength\itemsep{0cm}
    \item {\fontfamily{cmvtt}\selectfont  As shown in Figure 4G, overexpression of BCL-xL or, to a slightly lesser extent, BCL-xL-Mut1 inhibited the synergistic induction of apoptosis in [MASK] cells induced by the combined treatment with [MASK] and [MASK]. }
    \item {\fontfamily{cmvtt}\selectfont  We observed synergistic effects on [MASK] cells as well as three additional gastric cancer cell lines with FGFR2 amplification when [MASK] was combined with small molecular inhibitors Cpd22 and [MASK] targeting ILK and EGFR/HER2, respectively. }
    \item {\fontfamily{cmvtt}\selectfont  At most concentrations tested, [MASK] in combination with [MASK] synergistically inhibited the growth proliferation of [MASK] cells as shown by a Combination Index value < 1. }
    \item {\fontfamily{cmvtt}\selectfont  Combination treatment with [MASK] and [MASK] caused synergistically increased cell death in [MASK] and [MASK] cells. }
    \item {\fontfamily{cmvtt}\selectfont  In this study, [MASK] and [MASK] combination treatments had synergistic growth inhibitory effects on [MASK] cells, which were associated with a reduction in levels of phosphorylated Akt. }
    \item {\fontfamily{cmvtt}\selectfont  We further confirmed that the concomitant use of [MASK] and [MASK] has synergistic inhibitory effects on the growth of xenograft tumors derived from [MASK] cells. }
    \item {\fontfamily{cmvtt}\selectfont  Aneu-MKN45 developed a resistance to [MASK] which could be reversed by HZ08; Flow cytometry and western-blot indicates that HZ08-combination could induce apoptosis and increase the expression of apoptosis-related biomarkers on aneu-MKN45; in vivo study also reflect the same correlation between aneuploidy and cisplatin-resistance, which could be antagonized by HZ08 combination; When investigating the involved pathway, in anue-MKN45, the expression of molecules in p53 pathway was decreased; HZ08 could increase the expression of p53 down-stream molecules as well as elevate the activity of p53, while inhibiting Mdm2, the major negative regulator of p53; p53 inhibitor [MASK] could completely abrogate HZ08 's synergism effects, and mimic cisplatin-resistance on dip-MKN45.Lower p53 pathway expression that attenuates cisplatin-induced apoptosis might be at least partly the reason of cisplatin-resistance occurred in aneuploid [MASK] both in vitro and in vivo; Combination of HZ08 could sensitize cisplatin-induced apoptosis through the activation of the p53 pathway, therefore represented a synergism effect on aneuploid [MASK] cells. }
    \item {\fontfamily{cmvtt}\selectfont  We then investigated the effect of [MASK] in combination with [MASK] in cancer cell lines, and we demonstrated a synergistic growth inhibitory effect in GEO and [MASK] cells, evident also with suboptimal doses of [MASK]. }
    \item {\fontfamily{cmvtt}\selectfont  To explore the mechanism of synergistic effects by combining [MASK] and [MASK], we first detected apoptosis by PI staining in [MASK] and H1299 cells that displayed strong synergistic effects in the cytotoxicity assay. }
    \item {\fontfamily{cmvtt}\selectfont  The combination of 966 and either [MASK], [MASK], or [MASK] led to further reductions in cell growth than either agent alone in [MASK] cells, but these effects were additive, not synergistic. }
\end{itemize}

\subsubsection{Iteration 2}
\begin{itemize}[leftmargin=0.25cm]
    \setlength\itemsep{0cm}
    \item {\fontfamily{cmvtt}\selectfont  Preclinical [MASK] have indicated a synergistic effect for the [MASK] of anti-EGFR therapy plus [MASK] in breast cancer [MASK]. }
    \item {\fontfamily{cmvtt}\selectfont  As shown in Figure 4G, overexpression of BCL-xL or, to a slightly lesser extent, BCL-xL-Mut1 inhibited the synergistic induction of apoptosis in [MASK] [MASK] [MASK] by the combined [MASK] with [MASK] and [MASK]. }
    \item {\fontfamily{cmvtt}\selectfont  Another study presented that the [MASK] of [MASK] and WZB117 exerts a synergistic cytotoxic effect against breast cancer [MASK]. }
    \item {\fontfamily{cmvtt}\selectfont  In this study, we showed that clinically achievable concentrations of [MASK] and [MASK] in [MASK] could induce a synergistic growth arrest in several human PIK3CA wild-type cancer [MASK]. }
    \item {\fontfamily{cmvtt}\selectfont  The combinative [MASK] of FEN1 inhibitor and 1 nM [MASK] [MASK] a synthetic lethal effect, which synergistically suppressed cancer cell proliferation and significantly mediated apoptosis both in vitro and in vivo. }
    \item {\fontfamily{cmvtt}\selectfont  [MASK] of LDD1937 and AraC together showed a synergism in the cytotoxic effect on the [MASK] [MASK], and an additive effect was observed between [MASK] and LDD1937. }
    \item {\fontfamily{cmvtt}\selectfont  [MASK] using [MASK] in [MASK] with other chemotherapeutic drugs or natural compounds with anti-cancer potential may hold the key to identifying synergistic effects of the [MASK] therapy, thereby effectively decreasing the dosage required with better therapeutic efficiency when compared to usage as a monotherapy in the [MASK] of cancer. }
    \item {\fontfamily{cmvtt}\selectfont  Given that the [MASK] of [MASK] and IMiDs, including [MASK] and pomalidomide, have been shown to result in synergistic apoptotic MM cell death in vitro, the clinical activity of [MASK] could overcome IMiD [MASK] in myeloma patients, and our correlative data justify a phase [MASK] [MASK] of [MASK] and IMiD. }
    \item {\fontfamily{cmvtt}\selectfont  Interestingly, combinatory [MASK] with 50 muM [MASK] and 25 mM or 40 mM nicotinamide showed a synergistic effect on cell viability, which was observed in both cell [MASK]. }
    \item {\fontfamily{cmvtt}\selectfont  Chou and Talalay analysis of the data suggested that [MASK] of [MASK] and [MASK] was synergistic at a number of selected drug ratios and over a broad range of effective doses. }
\end{itemize}

\subsubsection{Iteration 3}
\begin{itemize}[leftmargin=0.25cm]
    \setlength\itemsep{0cm}
    \item {\fontfamily{cmvtt}\selectfont  The antioxidant capacity of AXT is ~1.5X that of vitamin E. This [MASK] reports on previously unknown findings concerning the synergistic antioxidative effects of combining AXT and HupA, using a previously established [MASK] system to characterize therapeutic agents that can scavenge [MASK] radicals and protect cells from tert-butyl hydroperoxide. }
    \item {\fontfamily{cmvtt}\selectfont  The resulting [MASK] index theorem of Chou-Talalay was [MASK] to calculate quantitatively whether the [MASK] of drugs [MASK] in an additive, synergistic, or antagonistic effect. }
    \item {\fontfamily{cmvtt}\selectfont  Recent in vitro [MASK] demonstrate that the [MASK] of H101 and [MASK] exerts a synergistic antitumor effect to uveal melanoma cells without enhanced toxicity to normal cells via a type of cell cycle block, reflecting H101 a promising agent in treating melanoma. }
    \item {\fontfamily{cmvtt}\selectfont  Interestingly, combinatory [MASK] with 50 muM [MASK] and 25 mM or 40 mM nicotinamide [MASK] a synergistic effect on cell viability, which was observed in both cell lines. }
    \item {\fontfamily{cmvtt}\selectfont  Given that the [MASK] of [MASK] and IMiDs, including [MASK] and pomalidomide, have been shown to result in synergistic apoptotic MM cell death in vitro, the [MASK] activity of [MASK] could overcome IMiD [MASK] in myeloma patients, and our correlative data justify a [MASK] [MASK] [MASK] of [MASK] and IMiD. }
    \item {\fontfamily{cmvtt}\selectfont  This technology offers a new [MASK] [MASK] pathway in reason of it 's targeted-specific pharmacodynamics and simplified pharmacokinetics that may improve the therapeutic effect towards tumor [MASK] In [MASK], here we have identified more than thirteen instances of synergism with other drugs that could improve therapeutic regimens, reducing toxicity and side effects. }
    \item {\fontfamily{cmvtt}\selectfont  Compared with the respective single anticancer action, PDA-NOC-ZnPc12+ nanoparticles [MASK] better anticancer efficacy in tumor-bearing mice, demonstrating the synergistic effect of [MASK] [MASK] with a cell cycle inhibitor and photosensitizer. }
    \item {\fontfamily{cmvtt}\selectfont  Gong et al., using multiple human PC cells, found that the [MASK] [MASK] with nexrutine and [MASK] [MASK] in significant alterations of proteins in the STAT3/NF-kappaB signaling axis, and growth [MASK] in a synergistic manner. }
    \item {\fontfamily{cmvtt}\selectfont  The results of the checkerboard [MASK] [MASK] that when colistin was [MASK] in [MASK] with PFK-158, synergistic activity was observed against the 31 colistin-resistant GNB. }
    \item {\fontfamily{cmvtt}\selectfont  The Bliss independence method was [MASK] to evaluate synergism between varying [MASK] of [MASK] and [MASK] in sNF96.2 cells, with the highest synergy attained at low doses for each [MASK] in cell proliferation [MASK]. }
\end{itemize}

  

\subsection{Dataset Access}

The original GDSC-combo dataset can be accessed at \url{https://figshare.com/articles/dataset/Original_screen_drug_combination_data/16843597}. 

LINCS dataset can be accessed at \url{https://lincsproject.org/LINCS/}.

GDSC dataset can be accessed at  \url{https://www.cancerrxgene.org/}.

CCLE dataset can be accessed at  \url{https://sites.broadinstitute.org/ccle/}.

NCI-60 dataset can be accessed at  \url{https://dtp.cancer.gov/discovery_development/nci-60/cell_list.htm}.

All datasets used in this paper are publicly available, and they are used for research purpose only in this paper. Further use of the datasets is subject to their own licenses. The derivated dataset on the GDSC-combo dataset is subject to the original license. All other data can be used for research only.

\end{document}